\documentclass[11pt,a4paper]{article}
\usepackage[hyperref]{acl2021}

\usepackage{times}
\usepackage{latexsym}
\usepackage{times}
\usepackage{multirow}
\usepackage{multicol}
\usepackage{graphicx}
\usepackage{amsmath}
\usepackage{amssymb}
\usepackage{bbm}
\usepackage{subfigure}
\usepackage{amsfonts}
\usepackage{bm}
\usepackage{bbding}
\usepackage{booktabs}
\usepackage{color}
\usepackage[T1]{fontenc}

\usepackage[utf8]{inputenc}
\usepackage{microtype}

\aclfinalcopy 

\title{Template-Based Named Entity Recognition Using BART}

\author{
  Leyang Cui$^{\dag \Diamond \heartsuit}$, Yu Wu$^\ddag$, Jian Liu$^{\Diamond \heartsuit}$, Sen Yang$^{\Diamond \heartsuit}$, Yue Zhang$^{\Diamond \heartsuit}$ \thanks{\ \  Corresponding Author} \\
  $^\dag$Zhejiang University \\
  $^\ddag$Microsoft Research Asia \\
  $^\Diamond$School of Engineering, Westlake University \\
  $^\heartsuit$Institute of Advanced Technology, Westlake Institute of Advanced Study \\
  \{cuileyang,liujian,yangsen,zhangyue\}@westlake.edu.cn\  
  Wu.Yu@microsoft.com \\
  }

\date{}

\begin{document}
\maketitle
\begin{abstract}
There is a recent interest in investigating few-shot NER, where the low-resource target domain has different label sets compared with a resource-rich source domain. Existing methods use a similarity-based metric. However, they cannot make full use of knowledge transfer in NER model parameters.
To address the issue, we propose a template-based method for NER, treating NER as a language model ranking problem in a sequence-to-sequence framework, 
where original sentences and statement templates filled by candidate named entity span are regarded as the source sequence and the target sequence, respectively. 
For inference, the model is required to classify each candidate span based on the corresponding template scores.
Our experiments demonstrate that the proposed method achieves 92.55$\%$ F1 score on the CoNLL03 (rich-resource task), and significantly better than fine-tuning BERT 10.88$\%$, 15.34$\%$, and 11.73$\%$ F1 score on the MIT Movie, the MIT Restaurant, and the ATIS (low-resource task), respectively.

\end{abstract}

\section{Introduction}
\begin{figure*}[t!]
    \centering
            \subfigure[Traditional sequence labeling method.]{
    \includegraphics[width=0.4\textwidth]{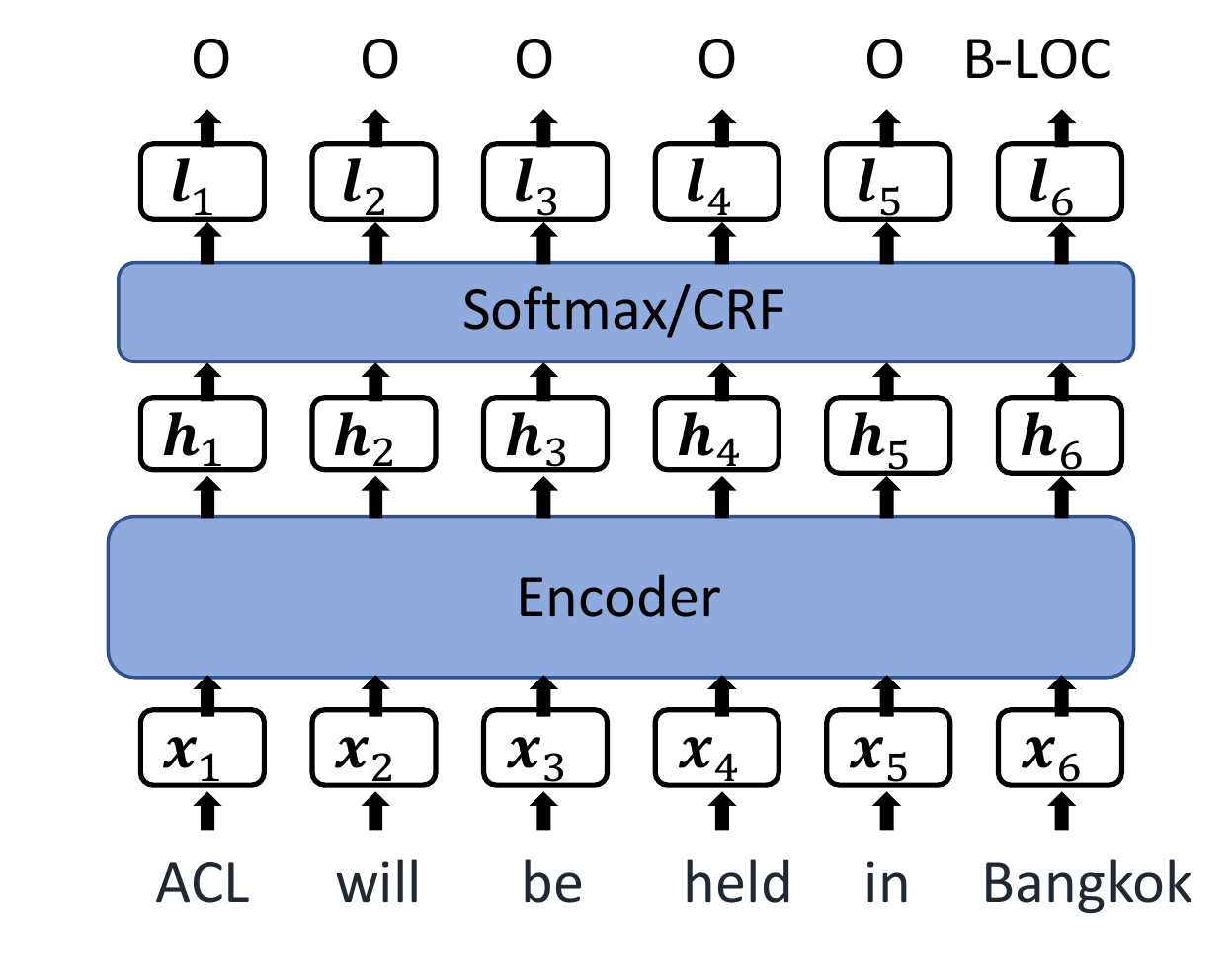}
    \label{fig:tradition}
    }
            \subfigure[Inference of template-based method.]{
    \includegraphics[width=0.33\textwidth]{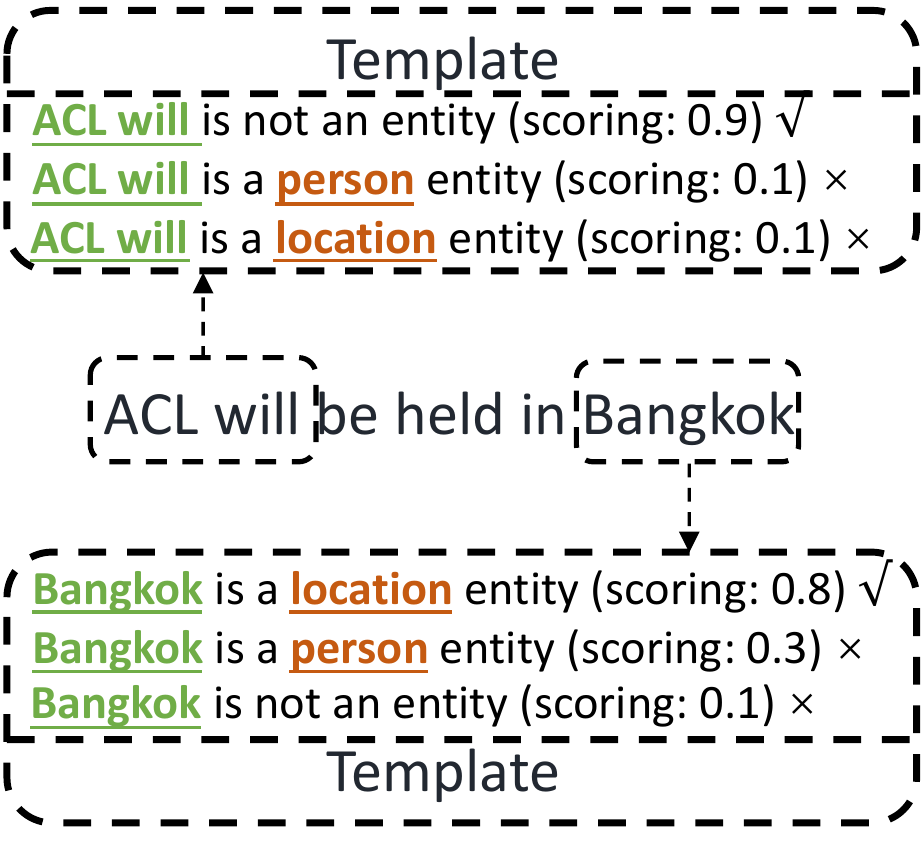}
    \label{fig:temp-test}
    }
        \subfigure[Training of template-based method. The template we use here is ``$\langle x_{i:j} \rangle$ is a $\langle y_k \rangle$ entity".]{
    \includegraphics[width=0.8\textwidth]{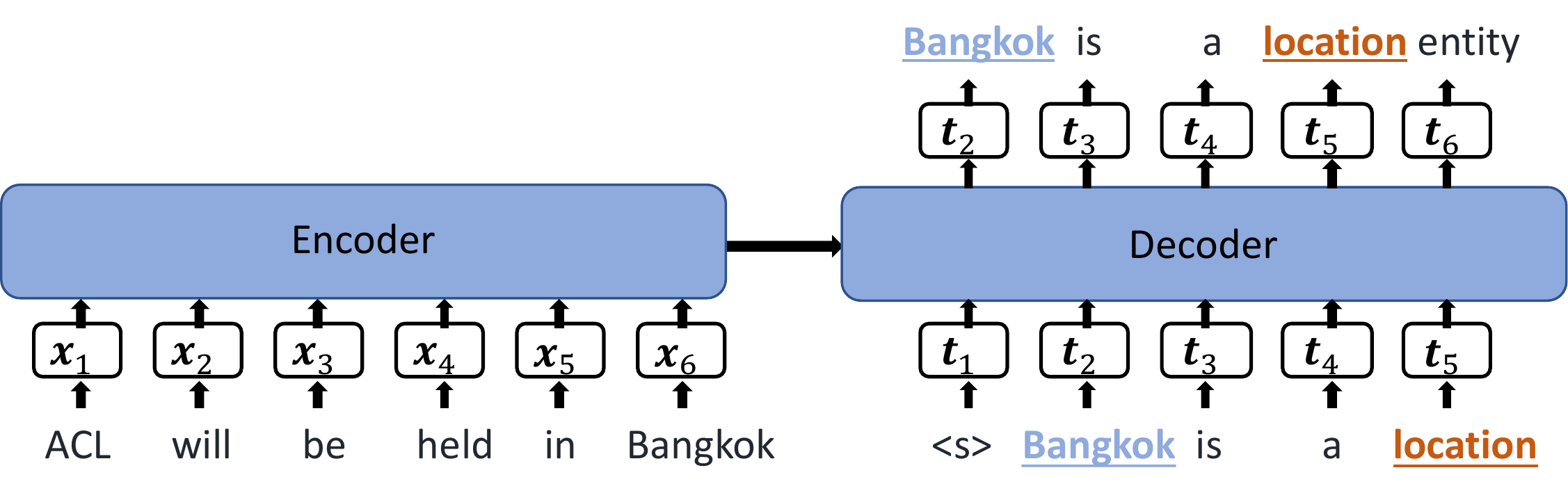}
    \label{fig:temp-train}
    }
    \caption{Overview of NER methods. \label{fig:intro}}
\end{figure*}

Named entity recognition (NER) is a fundamental task in natural language processing, which identifies mention spans from text inputs according to pre-defined entity categories \cite{conll03}, such as location, person, organization, etc. The current dominant methods use a sequential neural network such as BiLSTM \cite{lstm} and BERT \cite{bert} is used to represent the input text, and softmax \cite{lstm-cnn,dilated-convolutions,lan} or CRF \cite{lstm-crf,lstm-cnn-crf,luo2019hierarchical} output layers to assign named entity tags (e.g. organization, person and location) or non-entity tags on each input token. Such a system is illustrated in Figure~\ref{fig:tradition}.

\begin{figure}[t!]
    \centering
    \includegraphics[width=0.5\textwidth]{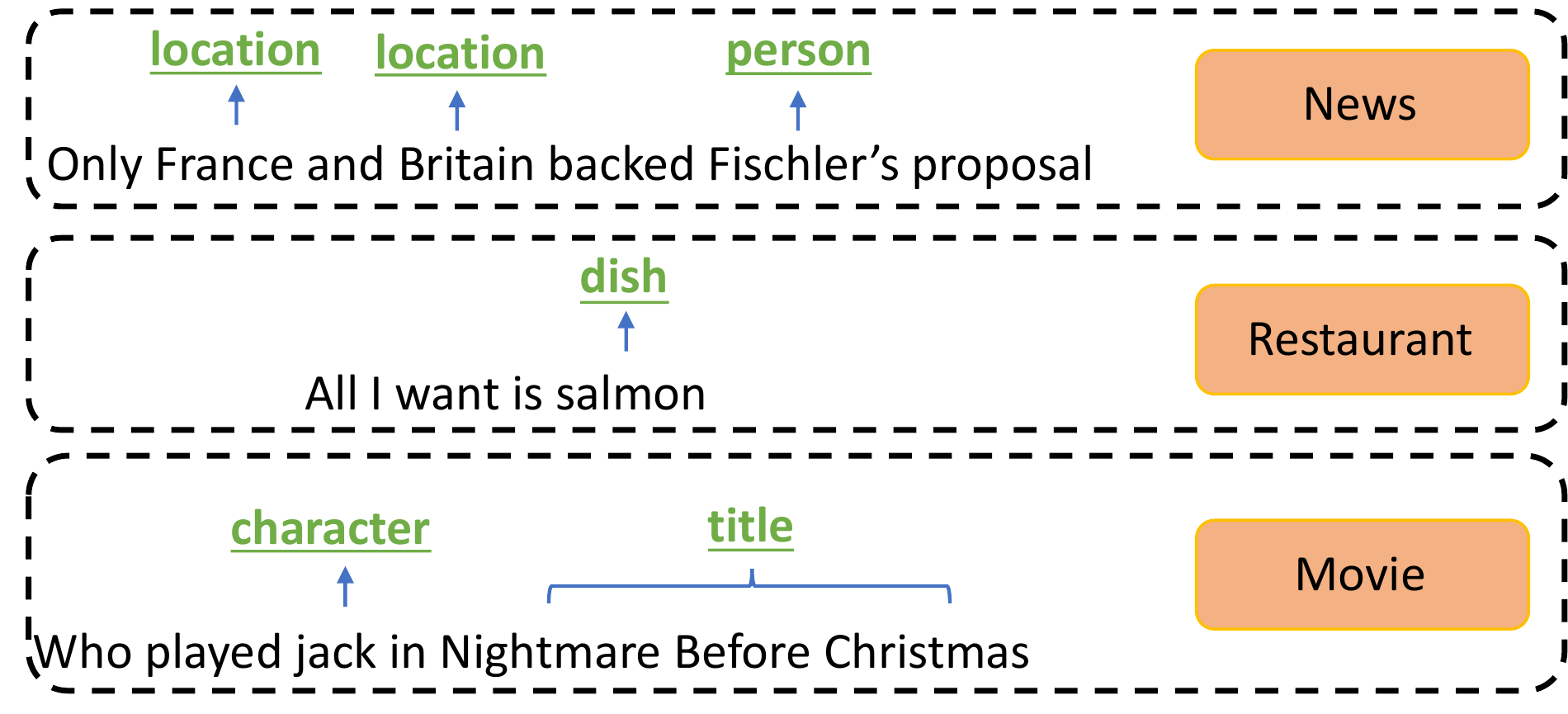}
    \caption{Example of NER on different domains.}
    \label{fig:example}
\end{figure}


Neural NER models require large labeled training data, which can be available for certain domains such as news, but scarce in most other domains. 
Ideally, it would be desirable to transfer knowledge from the resource-rich news domain so that a model can be used in target domains based on a few labeled instances. 
In practice, however, a challenge is that entity categories can be different across different domains. As shown in Figure~\ref{fig:example}, the system is required to identify {\it location} and {\it person} in the news domain, but {\it character} and {\it title} in the movie domain.
Both a softmax layer and CRF layer require a consistent label set between training and testing. 
As a result, given a new target domain, the output layer needs adjustment and training must be conducted again using both source and target domain, which can be costly.


A recent line of work investigates the setting of few-shot NER by using distance metrics \cite{label-agnostic,nearest-neighbor-crf, example-ner}.
The main idea is to train a similarity function based on  instances in the source domain, and then make use of the similarity function in the target domain as a nearest neighbor criterion for few-shot NER.

Compared with traditional methods, distance-based methods largely reduce the domain adaptation cost, especially for scenarios where the number of target domains is large. 
Their performance under standard in-domain settings, however, is relatively weak. 
In addition, their domain adaptation power is also limited in two aspects.
First, labeled instances in the target domain are used to find the best hyper-parameter settings for heuristic nearest neighbor search, but are not for updating the network parameters of the NER model. 
While being less costly, these methods cannot improve the neural representation for cross-domain instances. 
Second, these methods rely on similar textual patterns between the source domain and the target domain. This strong assumption may hinder the model performance when the target-domain writing style is different from the source domain. 


To address these issues, we investigate a template-based method for exploiting the few-shot learning potential of generative pre-trained language models to sequence labeling. Specifically, as shown in Figure~\ref{fig:intro}, BART \cite{bart} is fine-tuned with pre-defined templates filled by corresponding labeled entities. 
For example, we can define templates such as ``\textit{$\langle{\tt candidate\_span}\rangle$ is a $\langle{\tt entity\_type}\rangle$ entity}'', where $\langle{\tt entity\_type}\rangle$ can be ``\textit{person}'' and ``\textit{location}'', etc. Given the sentence ``\textit{ACL will be held in Bangkok}'', where ``\textit{Bangkok}'' has a gold label ``\textit{location}'', we can train BART using a filled template ``\textit{Bangkok is a location entity}'' as the decoder output for the input sentence.
In terms of non-entity spans, we use a template ``\textit{$\langle {\tt candidate\_span} \rangle$ is not a named entity}'', so that negative output sequences can also be sampled. During inference, we enumerate all possible text spans in the input sentence as named entity candidates, classifying them into entities or non-entities based on BART scores on templates. 

The proposed method has three advantages. First, due to the good generalization ability of pre-trained models \cite{gpt3, gao2020making}, the network can effectively leverage labeled instances in the new domain for tine-tuning. Second, compared with distance-based methods, our method is more robust even if the target domain and source domain have a large gap in writing style. 
Third, compared with traditional methods (pre-trained model with a softmax/CRF), our method can be applied to arbitrary new categories of named entities without changing the output layer, and therefore allows continual learning \cite{cl1}.

We conduct experiments in both resource-rich and few-shot settings. Results show that our methods give competitive results with state-of-the-art label-dependent approaches on the news dataset CoNLL03 \cite{conll03}, and significantly outperforms \citet{label-agnostic}, \citet{example-ner} and \citet{huang2020fewshot} when it comes to few-shot settings. 
To the best of our knowledge, we are the first to employ a generative pre-trained language model to address a few-shot sequence labeling problem. We release our code at \url{https://github.com/Nealcly/templateNER}.

\section{Related Work}

Neural methods have given competitive performance in NER. 
Some methods \cite{lstm-cnn, dilated-convolutions} treat NER as a local classification problem at each input token, while other methods use CRF \cite{lstm-cnn-crf} or a sequence-to-sequence framework \cite{tag-dependencies, gcdt}. \citet{lan} and \citet{refine} use a label attention network and Bayesian neural networks, respectively. \citet{luke} use entity-aware pre-training and obtain state-of-the-art results on NER. These approaches are similar to ours in the sense that parameters can be tuned in supervised learning, but unlike our method, they are designed for prescribed named entity types, which makes their domain adaptation costly with new few-shot entity types.

Our work is motivated by distance-based few-shot NER, which aims to minimize domain-adaptation cost. \citet{label-agnostic} copy the token-level label from nearest neighbors by retrieving a list of labeled sentences. \citet{nearest-neighbor-crf} improve \citet{label-agnostic} by using a Viterbi decoder to capture label dependencies estimated from the source domain.
\citet{example-ner} follow a two-step approach \cite{nest1,nest2}, which first detects spans boundary and then recognizes entity types by comparing the similarity with the labeled instance. While not updating the network parameters for NER, these methods rely on similar name entity patterns between the source domain and the target domain. One exception is \citet{huang2020fewshot}, who investigate noisy supervised pre-training and self-training method by using external noisy web NER data. Compared to their method, our method does not rely on self training on external data, yet yields better results.

There is a line of work using templates to solve natural language understanding tasks. The basic idea is to leverage information from pre-trained models, by defining specific sentence templates in a language modeling task. \citet{gpt3} first use prompt for few-shot learning in text classification tasks. \citet{schick2020exploiting} rephrase inputs as cloze questions for text classification. \citet{schick-etal-2020-automatically} and \citet{gao2020making} extend \citet{schick2020exploiting} by automatically generating label words and templates, respectively.
\citet{lama} extract relation between entities from BERT by constructing cloze-style templates.
\citet{auxiliary-absa} use templates to construct auxiliary sentences, and transform aspect sentiment task as a sentence-pair classification task. 
Our work is in line with exploiting pre-trained language model for templates-based NLP. 
While previous work considers sentence-level task as masked language modeling or uses language models to score a whole sentence, our method uses a language model to assign a score for each span given an input sentence.
To our knowledge, we are the first to apply template-based method to sequence labeling.

    
\section{Background}
We give the formal definition of few shot named entity recognition in Section~\ref{sec:fewshot} and traditional sequence labeling methods in Section~\ref{sec:sl}.

\subsection{Few shot Named Entity Recognition}
\label{sec:fewshot}
Suppose that we have a rich-resource NER dataset $\mathbb{H}= \{(\mathbf{X}_1^H,\mathbf{L}_1^H), ..., (\mathbf{X}_I^H,\mathbf{L}_I^H)\}$, where $\mathbf{X}^H=\{x_1^H,\dots,x_n^H\}$ is a sentence and $\mathbf{L}^H=\{l_1^H,\dots,l_n^H\}$ is its corresponding label sequence. We use $\mathcal{V}^H$ to denote the label set of the rich-resource dataset ($\forall l_i^H, l_i^H \in \mathcal{V}^H$). In addition, we have a low-resource NER dataset,  $\mathbb{L} = \{(\mathbf{X}_1^L,\mathbf{Y}_1^L), ..., (\mathbf{X}_J^L,\mathbf{Y}_J^L)\}$, and the number of its labelled sequence pairs is quite limited compared with the rich-resource NER dataset (i.e., $J \ll I$). Regarding the low-resource domain,  the target label vocabulary  $\mathcal{V}^L$ ($\forall l_i^L,l_i^L \in \mathcal{V}^L$) might be different from $\mathcal{V}^H$ (Figure~\ref{fig:example}). Our goal is to train an accurate and robust NER model with  $\mathbb{L}$ and $\mathbb{H}$ for the low-resource domain.


\subsection{Traditional Sequence Labeling Methods.}
\label{sec:sl}
Traditional methods (Figure~\ref{fig:tradition}) regard NER as a sequence labeling problem, where each output label consists of a sequence segmentation component $B$ (beginning of an entity), I (internal word in an entity), O (not an entity), and an entity type tag such as ``{\it person}'' and ``{\it location}''. For example, the tag ``{\it B-person}'' indicates the first word in a person type entity and the tag ``{\it I-location}'' indicates a token of a location entity not at the beginning. 
Formally, given $x_{1:n}$, the sequence labeling method calculates
\begin{equation}
\small
\begin{split}
       \mathbf{h}_{1:n} &= \textsc{Encoder}(x_{1:n}) \\
       p(\hat{l}_c) &= \textsc{softmax}(\mathbf{h}_c \mathbf{W}_{\mathcal{V}^R} + \mathbf{b}_{\mathcal{V}^R}) \ (c \in [1,...,n])
\end{split}
\label{eq:bert_output}
\end{equation}
where $d_h$ is the hidden dimension of the encoder, $\mathbf{W}_{\mathcal{V}^R} \in \mathbb{R}^{d_h \times |\mathcal{V}^R|}$ and $\mathbf{b}_{\mathcal{V}^R} \in \mathbb{R}^{|\mathcal{V}^R|}$ are trainable parameters, and $\hat{l}_c$ is the label estimation for $x_c$. 
We use BERT \cite{bert} and BART \cite{bart} as our \textsc{Encoder} to learn the sequence representation. 

A standard method for NER domain adaptation is to train a model using 
source-domain data $\mathbb{R}$ first, before further tuning the model using target domain instances $\mathbb{P}$, if available. 
However, since the label sets can be different, and consequently the output layer parameters ($\mathbf{W}_{\mathcal{V}^R} \in \mathbb{R}^{d_h \times |\mathcal{V}^R|}$, $\mathbf{b}_{\mathcal{V}^R} \in \mathbb{R}^{|\mathcal{V}^R|}$ and $\mathbf{W}_{\mathcal{V}^P} \in \mathbb{R}^{d_h \times |\mathcal{V}^P|}$, $\mathbf{b}_{\mathcal{V}^P} \in \mathbb{R}^{|\mathcal{V}^P|}$) can be different across domains. 
We train $\mathbf{W}_{\mathcal{V}^P}$ and $\mathbf{b}_{\mathcal{V}^P}$ from scratch using $\mathbb{P}$. However, this method does not fully exploit label associations (e.g., the association between ``{\it person}'' and ``{\it character}''), nor can it be directly used for zero-shot cases, where no labeled data in the target domain is available.

\section{Template-Based Method}
We consider NER as a language model ranking problem under a seq2seq framework. The source sequence of the model is an input text $\mathbf{X}=\{x_1,\dots,x_n\}$ and the target sequence $\mathbf{T}_{y_k,x_{i:j}}=\{t_1,\dots,t_m\}$ is a template filled by candidate text span $x_{i:j}$ and the entity type $y_k$. 
We first introduce how to create templates in Section~\ref{sec:template}, and then show the inference and training details in Section~\ref{sec:inference} and Section~\ref{sec:training}, respectively.
\subsection{Template Creation}
\label{sec:template}
We manually create the template, which has one slot for ${\tt candidate\_span}$ and another slot for the ${\tt entity\_type}$ label. We set a one to one mapping function to transfer the label set $\mathbf{L}=\{l_1,\dots,l_{|L|}\}$ (e.g., $l_k$=``LOC'') to a natural word set $\mathbf{Y}=\{y_1,\dots,y_{|L|}\}$ (e.g. $y_k$=``{\it location}''), and use words to define templates $\mathbf{T}_{y_k}^+$ (e.g. $\langle {\tt candidate\_span} \rangle$ is a location entity.). In addition, we create a non-entity template $\mathbf{T}^-$ for none of the named entity (e.g., $\langle{\tt candidate\_span}\rangle$ is not a named entity.). This way, we can obtain a list of templates $\mathbf{T} = [\mathbf{T}_{y_1}^+,\dots,\mathbf{T}_{y_{|L|}}^+, \mathbf{T}^-]$.
In Figure \ref{fig:temp-train}, the template $\mathbf{T}_{y_k, x_{i:j}}$ is ``$ \langle x_{i:j} \rangle$ is a $\langle y_k \rangle$'' and $\mathbf{T}^-_{x_{i:j}}$ is ``$\langle  x_{i:j} \rangle$ is not a named entity'', where $x_{i:j}$ is a candidate text span. 
\subsection{Inference}
\label{sec:inference}
We first enumerate all possible spans in the sentence $\{x_1,\dots,x_n\}$ and fill them in the prepared templates. For efficiency, we restrict the number of $n$-grams for a span from one to eight, so $8n$ templates are created for each sentence. Then, we use the fine-tuned pre-trained generative language model to assign a score for each template $\mathbf{T}_{y_k,x_{i:j}} =\{t_1,\ldots,t_m \}$, formulated as
\begin{equation}
\small
\begin{split}
       f(\mathbf{T}_{y_k,x_{i:j}}) = \sum_{c=1}^m \log  p(t_c|t_{1:c-1},\mathbf{X})
\end{split}
\end{equation}  

We calculate a score $f(\mathbf{T}_{y_k,x_{i:j}}^+)$ for each entity type and $f(\mathbf{T}^-_{x_{i:j}})$ for the none entity type by employing any pre-trained generative language model to score templates. Then we assign $x_{i:j}$ the entity type with the largest score to the text span. In this paper, we take BART as the pre-trained generative language models.


Our datasets do not contain nested entities. If two spans have text overlap and are assigned different labels in the inference, we choose the span with higher score as the final decision to avoid possible prediction contradictions. For instance, given the sentence ``{\it ACL will be held in Bangkok}'', the $n-$gram ``{\it in Bangkok}'' and ``{\it Bangkok}'' can be labeled ``ORG'' and ``LOC'', respectively, by using local scoring function $f(\cdot)$. In this case, we compare $f(\mathbf{T}_{\text{ORG},\text{``in Bangkok''}}^+)$ and $f(\mathbf{T}_{\text{LOC},\text{``Bangkok''}}^+)$, and choose the label which has a larger score to make the global decision.




\begin{table*}[t!]
\small
    \centering
    \begin{tabular}{c|c|c}
    \hline
          Entity Template $\mathbf{T}^+$ & None-Entity Template $\mathbf{T}^-$ & Dev F1 \\
    \hline
         $\langle {\tt candidate\_span} \rangle$ is a $\langle {\tt entity\_type} \rangle$ entity & $\langle {\tt candidate\_span}\rangle$ is not a named entity & 95.27\\
         The entity type of $\langle {\tt candidate\_span}\rangle$ is $\langle {\tt entity\_type} \rangle$ & The entity type of $\langle {\tt candidate\_span}\rangle$ is none entity  & 95.15\\
         $\langle {\tt candidate\_span} \rangle$ belongs to $\langle {\tt entity\_type} \rangle$ category & $\langle {\tt candidate\_span}\rangle$ belongs to none category & 88.42\\
         $\langle {\tt candidate\_span}\rangle$ should be tagged as $\langle {\tt entity\_type} \rangle$ & $\langle {\tt candidate\_span}\rangle$ should tagged as none entity & 76.80 \\
    \hline
    \end{tabular}
    \caption{Resulting using different templates.}
    \label{tab:template}
\end{table*}

\subsection{Training}
\label{sec:training}
Gold entities are used to create template during training. Suppose that the entity type of $x_{i:j}$ is $y_k$. We fill the text span $x_{i:j}$ and the entity type $y_k$ into $\mathbf{T}^+$ to create a target sentence $\mathbf{T}_{y_k,x_{i:j}}^+$. Similarly, if the entity type of $x_{i:j}$ is a none entity text span, the target sentence $\mathbf{T}_{x_{i:j}}^-$ is obtained by filling $x_{i:j}$ into $\mathbf{T}^-$. We use all gold entities in the training set to construct $(\mathbf{X},\mathbf{T}^+)$ pairs, and additionally create negative samples $(\mathbf{X},\mathbf{T}^-)$ by randomly sampling non-entity text spans. The number of negative pairs is 1.5 times that of positive pairs. 

Given a sequence pair $(\mathbf{X},\mathbf{T})$, we feed the input $\mathbf{X}$ to the encoder of the BART, and then we obtain hidden representations of the sentence
\begin{equation}
\small
           \mathbf{h}^{enc} = \textsc{Encoder}(x_{1:n}) 
\end{equation}

At the $c$ th step of the decoder, $\mathbf{h}^{enc}$ and previous output tokens $t_{1:c-1}$ are then as inputs, yielding a representation using attention \cite{transforms}
\begin{equation}
\small
           \mathbf{h}^{dec}_c = \textsc{Decoder}(\mathbf{h}^{enc}, t_{1:c-1}) 
\end{equation} 

The conditional probability of the word $t_c$ is defined as:
\begin{equation}
\small
   p(t_c|t_{1:c-1},\mathbf{X}) =  \textsc{softmax}(\mathbf{h}^{dec}_c \mathbf{W}_{lm} + \mathbf{b}_{lm})
\label{eq:bart_output}
\end{equation}
where $\mathbf{W}_{lm} \in \mathbb{R}^{d_h \times |\mathcal{V}|}$ and $\mathbf{b}_{lm} \in \mathbb{R}^{|\mathcal{V}|}$. $|\mathcal{V}|$ represents the vocab size of pre-trained BART. The cross-entropy between the decoder’s output and the original template is used as the loss function. 
\begin{equation}
\small
   \mathcal{L} = - \sum_{c=1}^m \log p(t_c|t_{1,c-1}, \mathbf{X})
\end{equation}



\subsection{Transfer Learning}
Given a new domain $\mathbb{P}$ with few-shot instances, the label set $\mathbf{L}^P$ (Section~\ref{sec:template}) can be different from what has been used for training the NER model. We thus fill the templates with the new domain label set for both training and testing, with the rest of the model and algorithms unchanged. In particular, given a small amount of ($\mathbf{X}^P,\mathbf{T}^P$), we create sequence pairs with the method described above for the low-resource domain, and fine-tuning the NER model trained on the rich-source domain. This process has low cost, yet can effectively transfer label knowledge.
Because the output of our method is a natural sentence instead of specific labels, both resource-rich and low-resource label vocabulary are subset of the pre-trained language model vocabulary ($\mathcal{V}^R,\mathcal{V}^P \subsetneqq \mathcal{V}$). This allows our method to make use of label correlations such as ``{\it person}'' and ``{\it character}'', and ``{\it location}'' and ``{\it city}'', for enhancing the effect of transfer learning across domains.

\section{Experiments}
We compare template-based BART with several baselines on both resource-rich settings and few-shot settings. We use the CoNLL2003 \cite{conll03} as the  resource-rich dataset. Following \citet{example-ner} and \citet{huang2020fewshot}, we use MIT Movie Review \cite{mit-dataset}, MIT  Restaurant Review \cite{mit-dataset} and ATIS \cite{atis} as the cross-domain few-shot dataset. 
Regarding the cross-domain transfer, there are unseen entity types in the three target few-shot datasets. Details of our training details and dataset statistics are shown in Appendix.





\subsection{Template Influence}
There can be different templates for expressing the same meaning. For instance ``{\it $\langle{\tt candidate\_span}\rangle$ is a person}'' can also be expressed by ``{\it $\langle{\tt candidate\_span}\rangle$ belongs to the person category}''.
We investigate the impact of manual templates using the CoNLL03 development set. Table~\ref{tab:template} shows the performance impact of different choice of templates. For instance, ``{\it $\langle {\tt candidate\_span} \rangle$ should be tagged as $\langle {\tt entity\_type} \rangle$}'' and ``{\it $\langle {\tt candidate\_span} \rangle$ is a $\langle {\tt entity\_type} \rangle$ entity}'' give 76.80\% and 95.27\% F1 score, respectively, indicating the template is a key factor that influences the final performance. Based on the development results, we use the top performing template ``$\langle {\tt candidate\_span}\rangle$ is a $\langle {\tt entity\_type} \rangle$ entity" in our experiments.


\subsection{CoNLL03 Results}
\paragraph{Standard NER setting.} We first evaluate the performance under the standard NER setting on CoNLL03.
The results are shown in Table~\ref{table:mp_conll}, where state-of-the-art methods are also compared. 
In particular, the sequence labeling BERT gives a strong baseline, F1 score at 91.73\%. 
We can see that even though the template-based BART is designed for few-shot named entity recognition, it performs competitively in resource-rich setting as well. For instance, our method outperforms sequence labeling BERT by 1.80\% on recall, which shows that our method is more effective in identifying the named entity, but also selecting irrelevant span.
Noted that though both sequence labeling BART and template-based BART make use of BART decoder representations, their performances have a large gap, where the latter outperforms the former by absolutely 1.30\% on F1 score, demonstrating the effectiveness of the template-based method. The observation is consistent with that of \citet{bart}, which shows that BART is not the most competitive for sequence classification. This may result from the nature of its seq2seq-based denoising autoencoder training, which is different from masked language modeling for BERT. 

To explore if templates are complementary for each other, we train three models using the first three templates reported in Table~\ref{tab:template}, and adopt an entity-level voting method to ensemble these three models.
There is a 1.21\% precision increase using ensemble, which shows that different templates may capture different type of knowledge. 
Finally, our method achieves a 92.55 \% F1 score by leveraging three templates, which is highly competitive with the best reported score. For computational efficiency, we use a single model for the subsequent few-shot experiments.

\paragraph{In domain few-shot NER setting.} We construct a few-shot learning scenario on the CoNLL03, where the number of training instances for some specific categories is quite limited by down-sampling.  In particular, we set ``MISC'' and ``ORG'' as the resource-rich entities, and ``LOC'' and ``PER'' as the low-resource entities. We down-sample the CoNLL03 training set, yielding 3,806 training instances, which includes 3,925 ``ORG'', 1,423 ``MISRC'', 50 ``LOC'' and 50 ``PER''. Since the text style is consistent in rich-resource and low-resource entity categories, we call the scenario in domain few-shot NER. 

\begin{table}[t!]
\centering
\small
\begin{tabular}{c|c|c|c}
     \hline
    {\bf Traditional Models} & {\bf P} & {\bf R} & {\bf F} \\
    \hline
    \citet{yang-etal-2018-design} & - & - & 90.77 \\
    \citet{lstm-cnn-crf} & - & - & 91.21 \\
    \citet{refine} & - & - & 92.02 \\
    \citet{luke}* & - & - & 94.30 \\
     Sequence Labeling BERT & 91.93 & 91.54 & 91.73 \\
     Sequence Labeling BART & 89.60 & 91.63 & 90.60 \\
     \hline
         {\bf Few-shot Friendly Models} & {\bf P} & {\bf R} & {\bf F} \\
     \hline
     \citet{label-agnostic} & - & - & 89.94 \\
     Template BART & 90.51 & 93.34 & 91.90 \\
     multi-template BART & 91.72 & 93.40 & 92.55 \\
     \hline
\end{tabular}
\caption{Model performance on the CoNLL03 \label{table:mp_conll}.The original result of BERT \cite{bert} was not achieved with the current version of the library as discussed and reported by \citet{stanislawek-etal-2019-named}, \citet{akbik-etal-2019-pooled} and \citet{refine}. * indicates training on external data.}
\end{table}

\begin{table}[t]
\centering
\small
\begin{tabular}{c|c|c|c|c|c}
     \hline
    Models & PER & ORG & LOC* & MISC* & Overall \\
    \hline
    BERT & 75.71 & 77.59 & 60.72 & 60.39 & 69.62 \\
    Ours & 84.49 & 72.61 & 71.98 & 73.37 & 75.59 \\
     \hline
\end{tabular}
\caption{In-domain Few-shot performance on the CoNLL03. * indicates it is a few-shot entity type. \label{table:conll_few}}
\end{table}

\begin{table*}[t!]
\centering
\small
\begin{tabular}{c|c|c|c|c|c|c|c}\hline
\multicolumn{8}{c}{\textit{MIT Movie}}\\
\hline

    Source & Methods & 10 & 20 & 50 & 100 & 200 & 500 \\
\hline
    \multirow{2}{*}{None} & Sequence Labeling BERT & 25.2 & 42.2 & 49.64 & 50.7 & 59.3 & 74.4 \\
     & Template-based BART & 37.3 & 48.5 & 52.2 & 56.3 & 62.0 & 74.9 \\
\hline
    \multirow{6}{*}{CoNLL03} & \citet{label-agnostic} &\ 3.1 &\ 4.5 &\ 4.1 &\ 5.3 &\ 5.4 &\ 8.6 \\
    & \citet{example-ner} & 40.1 & 39.5 & 40.2 & 40.0 & 40.0 & 39.5 \\
    & \citet{huang2020fewshot}* & 36.4 & 36.8 & 38.0 & 38.2 & 35.4 & 38.3 \\
    & Sequence Labeling BERT & 28.3 & 45.2 & 50.0 & 52.4 & 60.7 & 76.8 \\
    & Sequence Labeling BART &13.6&30.4&47,8&49.1&55.8&66.9\\
    & Template-based BART & 42.4 & 54.2 & 59.6 & 65.3 & 69.6 & 80.3 \\
    \hline
    \multicolumn{8}{c}{\textit{MIT Restaurant}}\\ \hline
    \multirow{2}{*}{None} & Sequence Labeling BERT & 21.8 & 39.4 & 52.7 & 53.5 & 57.4 & 61.3 \\
    & Template-based BART& 46.0 & 57.1 & 58.7 & 60.1 & 62.8 & 65.0 \\
    \hline
    \multirow{6}{*}{CoNLL03} & \citet{label-agnostic} &\ 4.1 &\ 3.6 &\ 4.0 &\ 4.6 &\ 5.5 &\ 8.1 \\
    & \citet{example-ner} & 27.6 & 29.5 & 31.2 & 33.7 & 34
    .5 & 34.6 \\
       & \citet{huang2020fewshot} & 46.1 & 48.2 & 49.6 & 50.0 & 50.1 \\
     & Sequence Labeling BERT & 27.2 & 40.9 & 56.3 & 57.4 & 58.6 & 75.3 \\
    & Sequence Labeling BART &8.8&11.1&42.7&45.3&47.8&58.2\\
    & Template-based BART & 53.1 & 60.3 & 64.1 & 67.3 & 72.2 & 75.7 \\\hline
    \multicolumn{8}{c}{\textit{ATIS}}\\ \hline

    \multirow{2}{*}{None} & Sequence Labeling BERT & 44.1 & 76.7 & 90.7 & - & - & - \\
    & Template-based BART & 71.7 & 79.4 & 92.6 & - & - & - \\
\hline
    \multirow{6}{*}{CoNLL03} &\citet{label-agnostic} &\ 6.7 &\ 8.8 &\ 11.1 & - & - & -  \\
    &\citet{example-ner} & 17.4 & 19.8 & 22.2 & - & - & -  \\
    &\citet{huang2020fewshot} & 71.2 & 74.8 & 76.0 & - & - & -\\
    & Sequence Labeling BERT & 53.9 & 78.5 & 92.2 & - & - & - \\
    &  Sequence Labeling BART &51.3&74.4&89.9 & - & - & -\\
    & Template-based BART & 77.3 & 88.9 & 93.5 & - & - & - \\
\hline

\hline
\end{tabular}
\caption{\label{tab:movie}Cross-domain few-shot NER performance on different test sets. * indicates training on external data. 10 indicates 10 instances for each entity types.}
\vspace{-0.1cm}
\end{table*}

As shown in Table~\ref{table:conll_few}, sequence labeling BERT and template-based BART show similar performance in resource-rich entity types, while our method significantly outperforms BERT by 11.26 and 12.98 F1 score in ``LOC'' and ``MISC'', respectively. It demonstrates that our method has a stronger modeling capability for in-domain few-shot NER, and indicates that the proposed method can better transfer the knowledge between different entity categories. 


\subsection{Cross-domain Few-Shot NER Result}

We evaluate the model performance when the target entity types are different from the source-domain, and only a small amount of labeled data is available for training. We simulate the cross-domain low-resource data scenarios by random sampling training instances from a large training set as the training data in the target domain.  
We use different numbers of instances for training, randomly sampling a fixed number of instances per entity type (10, 20, 50, 100, 200, 500 instances per entity type for MIT Movie and MIT restaurant, and 10, 20, 50 instances per entity type for ATIS). If an entity has a smaller number of instances than the fixed number to sample, we use all of them for training. 
The results on few-shot experiments using MIT Movie, MIT Restaurant and ATIS are shown in Table~\ref{tab:movie}, where the methods of \citet{label-agnostic}, \citet{example-ner} and \citet{huang2020fewshot} are also compared. 

We first consider a training-from-scratch setting, where no source-domain data is used. Distance-based methods cannot suit this setting. Compared with the traditional sequence labeling BERT method, our method can make better use of few-shot data.
In particular, with as few as 20 instances per entity type, our method gives a F1 score of 57.1\%, higher than BERT using 100 instances per entity type on MIT Restaurant.

We further investigate how much knowledge can be transferred from the news domain (CoNLL03). In this setting, we further train the model which is trained on the news domain.
It can be seen from the Table~\ref{tab:movie} that on all the three datasets, the few-short learning methods outperform sequence labeling BERT and BART methods when the number of training instances is small. For example, when there are only 10 training instances, the method of \citet{example-ner} gives a F1 score of 40.1\% on MIT Movie, as compared to 28.3\% by BERT, despite that BERT requires re-training with a different output layer on both CoNLL03 and MIT Movie. However, as the number of training instances increase, the advantage of baseline few-shot methods decreases. When the number of instances grows as large as 500, BERT outperforms all existing methods. Our method is effective in both 10 instances and 500 instances, outperforming both BERT and baseline few-shot methods.

Compared with the distance-based method  \cite{label-agnostic,example-ner,huang2020fewshot}, our method shows more improvement when the number of target-domain labeled data increases, because the distance-based method just optimizes its searching threshold rather than updating its neural network parameters. We can see that the performance of distance-based methods remains the same as the labeled data increasing. For example, the performance of \citet{huang2020fewshot} increases only 1.9\% F1 score when the number of instances per entity type increase from 10 to 500.
Both BERT and our method perform better than training from scratch. Our model average increases 6.6, 6.9 and 5.4 F1 score on MIT restaurant, MIT movie and ATIS, respectively, which is significantly higher than 3.1, 1.9 and 4.3 F1 score in BERT. 
This shows that our model is more successful in transferring knowledge learned from the source domain. One possible explanation is that our model makes more use of the correlations between different entity type labels in the vocabulary as mentioned earlier, which BERT cannot achieve due to treating the output as discrete class labels.




\begin{figure*}[t!]
    \centering
        \subfigure[High Frequency.]{
    \includegraphics[width=0.3\textwidth]{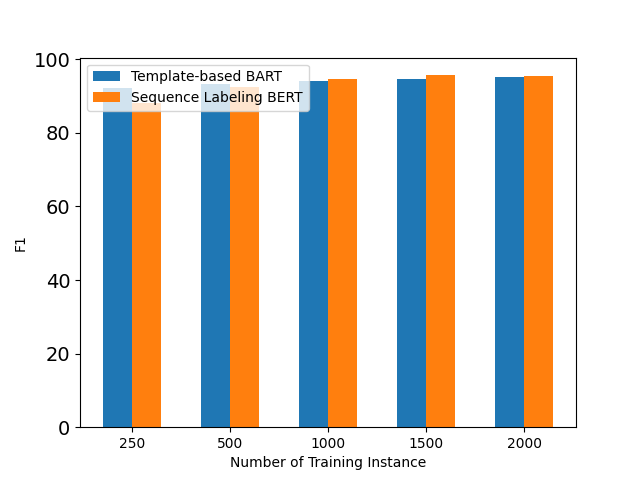}
    }
        \subfigure[Mid Frequency.]{
    \includegraphics[width=0.3\textwidth]{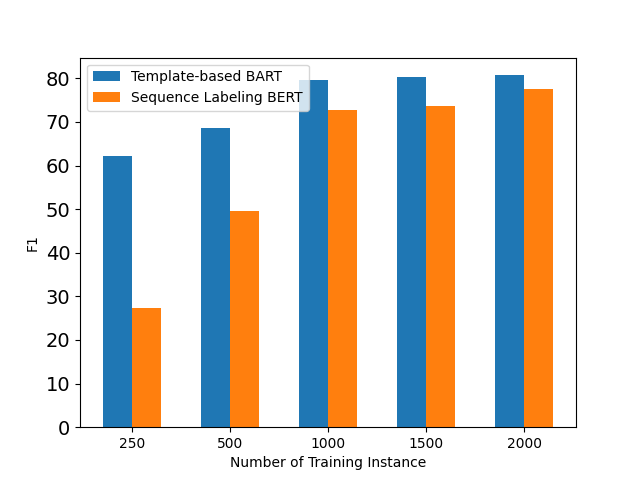}
    }
    \subfigure[Low Frequency.]{
    \includegraphics[width=0.3\textwidth]{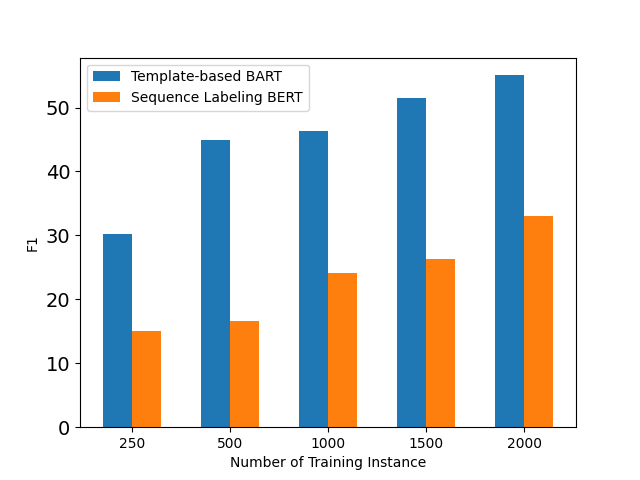}
    }
    \caption{Comparison of F1 with different frequency entity types on ATIS.}
    \label{fig:atis_number}
\end{figure*}

\begin{figure}
    \centering
    \includegraphics[width=0.45\textwidth]{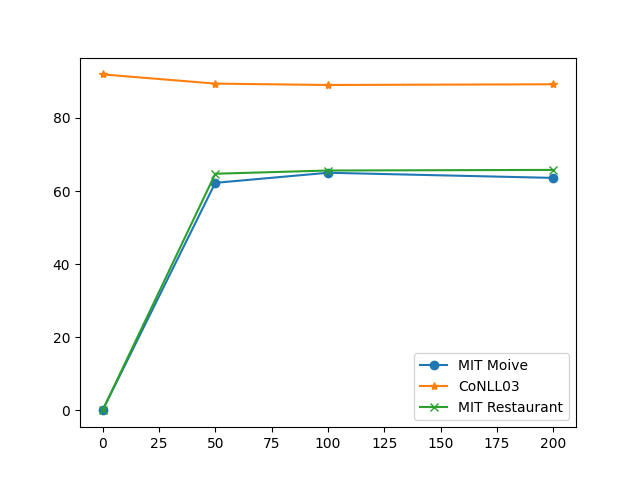}
    \caption{Continual learning experiments.}
    \label{fig:online}
\end{figure}

\begin{figure}[t!]
    \centering
        \subfigure[Sequence labeling BERT.]{
    \includegraphics[width=0.22\textwidth]{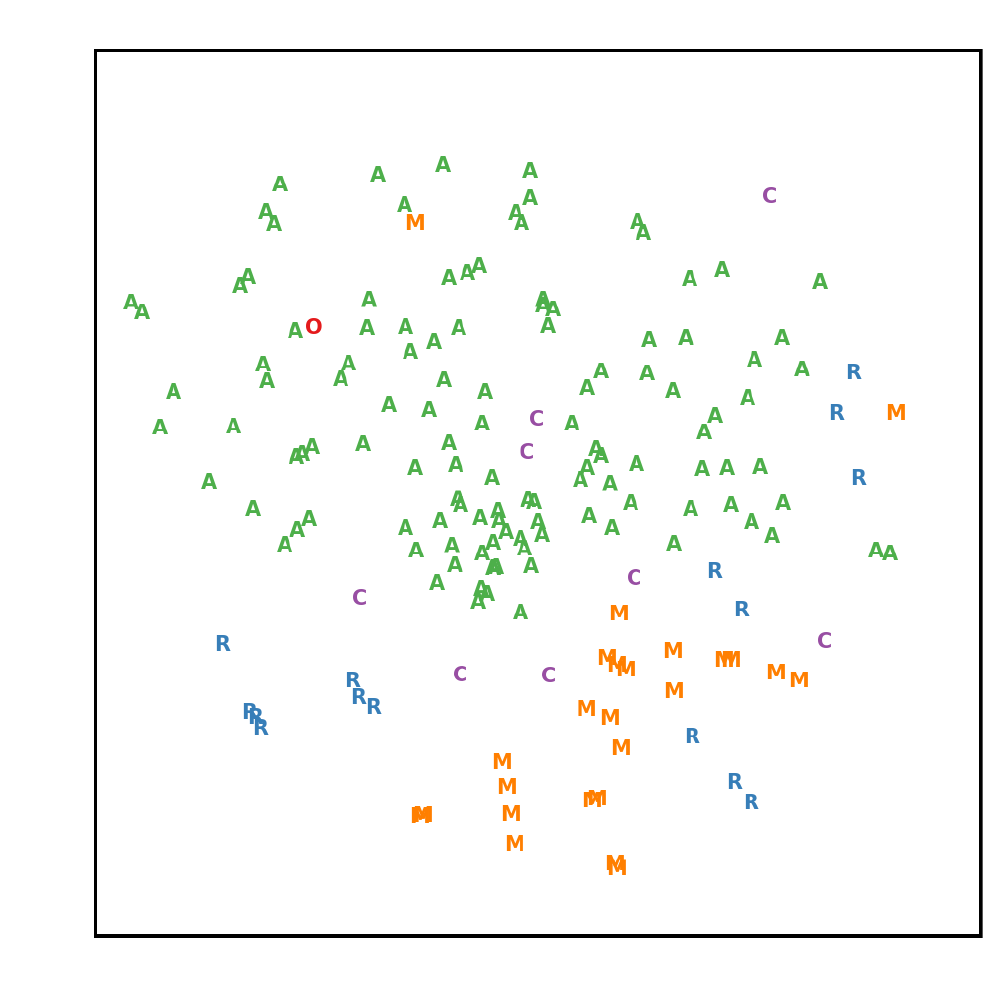}
    }
        \subfigure[Template-based BART.]{
    \includegraphics[width=0.22\textwidth]{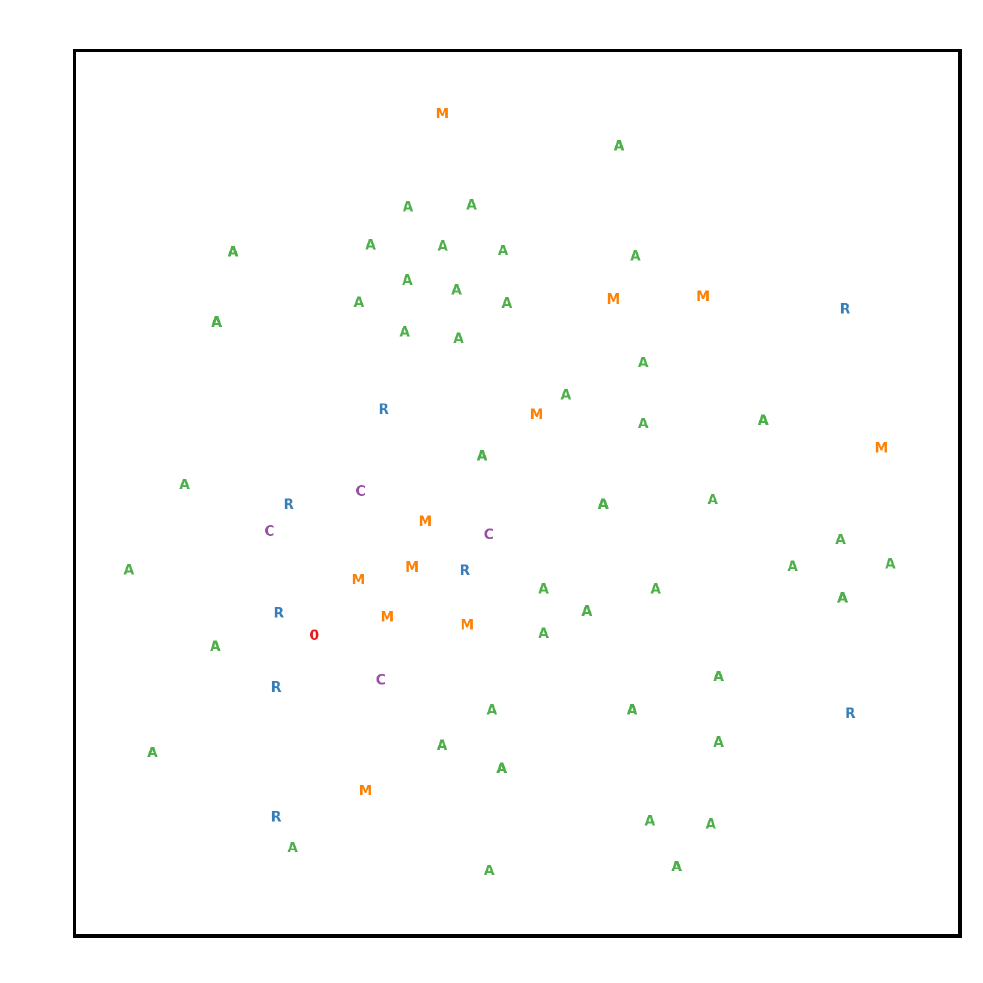}
    }
    \caption{Visualization of the output embedding. ``C''--CoNLL03, ``A''--ATIS, ``M''--MIT Movie, ``R''--MIT Restaurant.}
    \label{fig:vis}
\end{figure}


\subsection{Discussion}
\label{sec:discussion}


\paragraph{Impact of entity frequencies in training data.} To explore the relation between recognition accuracy and the  frequency of an entity type in training, we split ATIS test set into three subset based on the entity frequency in training. The most 33\% frequency entities are put into {\it high frequency} subset, the last 33\% frequency entities are put into {\it low frequency} subset, and the remaining are put into {\it mid frequency} subset. Figure~\ref{fig:atis_number} shows the F1 score of BERT and our method against the number of training instance in the three subsets. As the number of training instances increases, the performance of all models increases. Our method outperforms sequence labeling BERT by a large margin, especially on the {\it mid frequency} and {\it low frequency} subsets, which demonstrates that our method is more robust in few-shot settings. 

\paragraph{Continual Learning NER}
In continual learning setting \cite{cl1}, all the baselines that we have in Table~\ref{tab:movie} face limitations. The sequence labeling BERT method needs re-training using all training data each time a new entity type is encountered, which is highly expensive. The distance based methods cannot make use of all available data for improving representation learning. Figure~\ref{fig:online} shows performance of our method on MIT movie, MIT restaurant and CoNLL03, when we continue to train our CoNLL03 model on both MIT movie and MIT restaurant. The performance on the CoNLL03 only slightly decrease when we continue training the model on the MIT Movie and MIT Restaurant dataset, demonstrating the robustness of our method in the continual learning setting. 

\paragraph{Visualization.} We explore why our model works well in the low-resource domain by visualizing the output layer. We train BERT and our method on all four datasets, and use t-SNE \cite{tsne} to project the output layer into 2-dimensions, where the output layer for sequence labeling BERT and template-based BART are $\mathbf{W}_{\mathcal{V}^R}$ in Eq~\ref{eq:bert_output} and  $\mathbf{W}_{lm}$ in Eq~\ref{eq:bart_output}, respectively. In Figure~\ref{fig:vis}, each dot represents a row in the output matrix (corresponding to a label embedding).
We can see that output layer embeddings of BERT are clustered based on dataset while the vectors of template-based BART are sparsely distributed in the space. It indicates that our output matrix is more domain independent, and our method enjoys better generalization ability across different domains. 

\paragraph{Error Types}
We find that most mistakes are caused by the domain distance between high-resource data and low-source NER data. As shown in Fig~\ref{fig:vis}, Template-based methods rely on label semantics. If the embedding of the word with a few-shot labels is far from that with in-domain labels, the model shows lower performance on that label type. Taking 50 examples per entity type on MIT movie as an example, ``ACTOR'' is similar to ``PERSON'' in CoNLL03, and achieves 84.81 F1. The embedding of ``SONG'' is far from the existing labels in CoNLL03, and only achieves 34.97 F1. In contrast, sequence labeling BERT does not suffer from this distance, because BERT cannot draw label correlation between two domains, it achieves 53.98 and 40.13 on ``ACTOR'' and ``SONG'', respectively. 


\section{Conclusion}
We investigated template-based few-shot NER using BART as the backbone model. In contrast to the traditional sequence labeling methods, our method is more powerful on few-shot NER, since it can be fine-tuned for the target domain directly when new entity categories exist. Experiment results show that our model achieves competitive results on a rich-resource NER benchmark, and outperforms traditional sequence labeling methods and distance-based methods significantly on the cross-domain and few-shot NER benchmarks. 

\section*{Acknowledgments}
We thank all anonymous reviewers for their constructive comments.
This work is supported by National Science Foundation of China (Grant No. 61976180).

\bibliographystyle{acl_natbib}
\bibliography{custom}






\end{document}


\maketitle

\begin{appendix}
\section{Implementation Details}
We train BERT with a softmax classifier following \citet{bert}, updating parameters using Adam with an initial learning rate of 1e-5, and a batch size of 32. We employ a learning rate decay with 0.05 per iteration. We fine-tune BART with min-batches of size 64 using the Adam optimizer with a 2e-5 learning rate and a warmup period.

\section{Dataset}
We take the standard split of CoNLL03 by following \citet{conll03}, and splits MIT Movie Review, MIT Restaurant Review and ATIS by following \citet{mit-dataset}.
Table~\ref{tab:data_statistic} presents detailed statistics of our datasets.
The standard precision, recall and F1 score are used for model evaluation.

\begin{table}[h!]
    \centering
    \begin{tabular}{c|c|c|c}
    \hline
         Dataset & \# Train & \# Test & \# Entity \\
    \hline
    CoNLL03 & 12.7k & 3.2k & 4 \\
    MIT Restaurant & 7.6k & 1.5k & 8 \\
    MIT Review & 7.8k & 2k & 12 \\
    ATIS & 4.6k & 850 & 79 \\
    \hline
    \end{tabular}
    \caption{Statistic of datasets.}
    \label{tab:data_statistic}
\end{table}

\end{appendix}

\bibliography{anthology,custom}
\bibliographystyle{acl_natbib}


